\documentclass[a4paper,10pt,twocolumn]{article}
\usepackage[margin=1in]{geometry}
\usepackage{graphicx} 
\usepackage[style=ieee]{biblatex}
\usepackage{sectsty}
\usepackage{mathptmx}
\usepackage{amsmath}
\usepackage{amsfonts}
\usepackage{xcolor}
\usepackage{algorithm}
\usepackage{algpseudocode}
\usepackage{titling}
\usepackage{float}
\sectionfont{\large\MakeUppercase}
\usepackage[]{authblk}

\title{Demonstrating Onboard Inference for Earth Science Applications with Spectral Analysis Algorithms and Deep Learning}
\author[1]{Itai Zilberstein\thanks{itai.m.zilberstein@jpl.nasa.gov. \copyright 2024. All rights reserved.}}
\author[1]{Alberto Candela}
\author[1]{Steve Chien}
\author[2]{David Rijlaarsdam}
\author[2]{Tom Hendrix}
\author[2]{L\'eonie Buckley}
\author[2]{Aubrey Dunne}
\affil[1]{Jet Propulsion Laboratory, California Institute of Technology, United States}
\affil[2]{Ubotica Technologies, Ireland}

\makeatletter
\def\@maketitle{%
  \null
  \begin{center}%
    {\bf \Large \@title \par}%
    \vskip 1.5em%
    {\large
      \lineskip .5em%
      \begin{tabular}[t]{c}%
        \normalsize \@author \\
       
      \end{tabular}\par}%
  \end{center}%
  \par
  \vskip 1.5em}
\makeatother

\begin{document}

\maketitle

\begin{abstract}
In partnership with Ubotica Technologies, the Jet Propulsion Laboratory is demonstrating state-of-the-art data analysis onboard CogniSAT-6/HAMMER (CS-6). CS-6 is a satellite with a visible and near infrared range hyperspectral instrument and neural network acceleration hardware. Performing data analysis at the edge (e.g. onboard) can enable new Earth science measurements and responses. We will demonstrate data analysis and inference onboard CS-6 for numerous applications using deep learning and spectral analysis algorithms.

\end{abstract}

\section{Introduction}

The capabilities of in-orbit assets to perform Earth science has skyrocketed in recent years. New space providers are deploying satellites that possess state-of-the-art multi- and hyperspectral instruments along with processors that raise the ceiling for onboard computation. CogniSAT-6/HAMMER (CS-6) is one such spacecraft that has a visible and near infrared range hyperspectral instrument and AI acceleration hardware, enabling advanced edge data analysis \cite{cs6}. CS-6, which is a 6U CubeSat, launched in March of 2024. It is in a sun-synchronous orbit at an orbital height of around 500 km. Onboard CS-6 there is a Myriad X Vision Processing Unit (VPU) which can perform rapid computer vision, image signal processing, and neural network execution. CS-6 employs the HyperScape 100 instrument that can take hyperspectral measurements in the range $440\text{nm}{-}884$nm and achieves a ground sample distance of $5$ meters per pixel.

Analyzing data onboard serves several key functionalities including rapid response to detected phenomena and reducing data volume through identification of unusable data. The former relies on inferring science properties of a data acquisition to direct intelligent planning of future measurements.  Performing this computation at the edge (e.g. onboard) is key to enabling new, time-sensitive measurements of rare Earth phenomena that would otherwise be unobtainable if ground analysis was required. A spacecraft can self-cue and take another measurement in the same overflight using the knowledge from the first acquisition to direct the second one. This concept, called \textit{dynamic targeting}, would enable higher-resolution pinpoint measurements \cite{dt-igarss,dt-jais}. Alternatively, a spacecraft can cross-cue another spacecraft to obtain rapid follow-up imagery of key targets. Onboard data analysis realizes a key component of NASA’s New Observing Strategies (NOS) program which aims to advance observation systems.
\begin{figure}[t]
\centering
\includegraphics[width=\columnwidth]{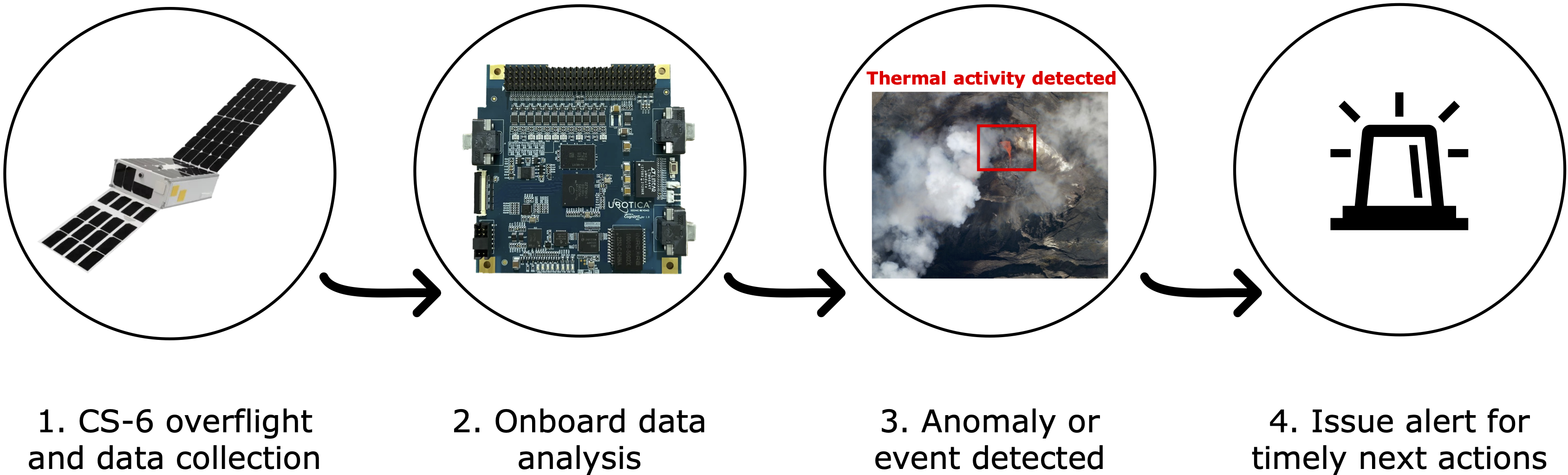}
\caption{Overview of a rapid response workflow for volcano monitoring. \textit{Image sources: Ubotica Technologies (steps 1 \& 2), Planet Labs (step 3)}.}
\label{fig:first_fig}
\end{figure}
Figure \ref{fig:first_fig} illustrates a rapid-response use-case for onboard data inference. Detection of thermal activity triggers communication to ground stations or other assets for quick follow-up actions. Inter-satellite links (ISL) enables satellites such as CS-6 to cross-cue other  sensing assets or participate in decentralized scheduling \cite{zilberstein-icaps-2024}.

In this work, we focus on the development and deployment of onboard algorithms, including spectral analysis algorithms and deep neural networks such as convolutional neural networks (CNNs), to perform inference for both image analysis and spectral signature detection in the visible and near infrared spectral ranges. Leveraging AI acceleration hardware to perform spectral analysis is novel and pioneers new avenues for edge computing.  We target numerous Earth science applications ranging from the detection of clouds, wildfires, volcanic activity, and harmful algal blooms to surface water, vegetation, and mineral mapping and land-use classification. This paper details the development of inference algorithms that will be demonstrated onboard CS-6 and other spacecraft. This ongoing effort consists of composing datasets for each application using automated labeling techniques, designing and implementing algorithms within the constraints of flight hardware including training machine learning models, validating inference and execution, and a flight demonstration.

\section{Methods}

\subsection{Datasets}

CS-6 data is limited due to its recent launch (March 2024). For spectral analysis, we leverage data from the USGS spectral library \cite{USGS}. The image datasets are composed of 190 scenes from the Menut satellite operated by Open Cosmos and several hundred scenes from Planetscope data. We use the red, blue, green, and near infrared bands from these data products. CS-6 bands are stretched to enhance the data and enable compatibility with models trained on other satellite scenes. Let $q_1$ and $q_{99}$ be the $1$st and $99$th quantiles of a band. We stretch (onboard) each pixel, $p$, in a band so that it is in the range $[v_{min}, v_{max}]$ using the formula
\small
\begin{equation}
\textsc{min}\left[ \textsc{max} \left(v_{min} +  \frac{v_{max} - v_{min}}{(q_{99} - q_1)}  \cdot (p - q_1), v_{min} \right), v_{max}\right]{.}
\end{equation}
\normalsize
We derive automated labeling techniques from the Haze Optimized Transform (HOT) method for clouds \cite{hot}, Normalized Difference Water Index (NDWI) method for surface water extent (SWE) \cite{ndwi}, band thresholds for thermal activity \cite{mason2023fully}, ESA's WorldCover maps for land use \cite{worldcover}, and Sentinel-2 imagery for algal blooms. 
Methods such as HOT and NDWI can be directly computed from VNIR data. We compute the NDWI mask using the equation
\begin{equation}
NDWI= \frac{Green - NIR}{Green + NIR}.
\end{equation}

The NDWI product is then thresholded using Otsu’s method to obtain a binary surface water mask for a scene. The HOT method leverages the relationship between blue and red band values for non-cloudy pixels. The red band is less affected by atmospheric haze, while the blue band has more scattering. The clear-sky line defines this correlation and points that fall close to this line signify clear pixels, while points far from this line signify cloud or haze interference. Note that Planetscope scenes are delivered with cloud masks, therefore we apply the HOT method to label Menut and CS-6 data products. We compute the clear-sky line by selecting the $0.15\%$ of data points with the smallest blue band value. These points are divided into 20 bins, and for each bin the 20 points with the highest red band value are selected. Given these 400 points, we fit a line using linear regression. Let $m$ and $b$ define the clear-sky line. We compute the HOT value as
\begin{equation}
HOT= | m \cdot Blue - Red |+ \frac{b}{\sqrt{1+ m^2}}.
\end{equation}

Like NDWI, the HOT product is thresholded using Otsu’s method to obtain a binary cloud mask. Despite using automated techniques, the labelled data is human-verified to ensure quality. 

Due to CS-6 having only visible and near infrared data products, the science return is limited for certain applications. Without a higher wavelength, cloud classification includes noise from snow, shorelines, and other high-reflectance objects. SWE has noise from shadows and urban areas. There are obvious limitations in thermal  detection without VSWIR or TIR sensing. Despite these drawbacks, the models perform with high accuracy as shown in the evaluation.

\subsection{Spectral Analysis Algorithms}

Spectral analysis is used for two different applications: mineral and vegetation mapping. We engineer these algorithms to leverage the AI acceleration hardware onboard CS-6, a novel approach to deploying spectral algorithms. 

We use three common methods: spectral angle mapper (SAM), matched filters (MF), and the Reed-Xiaoli anomaly detector (RX). 

SAM is a function that measures similarity between any two spectra: $x$ and $y$. It generalizes the notion of an angle between two vectors in N-space. Small SAM values indicate a high similarity, and vice versa. It is computed as follows:
\begin{equation}
SAM(x,y) = \cos ^{-1} \left ( \frac{x \cdot y}{||x||^2 ~ ||y||^2} \right).
\end{equation}

MF also quantifies similarity between two spectra, but it scales and normalizes the response by using scene statistics $\mu$ (mean) and $\Sigma$ (covariance). Larger MF values indicate a stronger match between spectrum $x$ and a target of interest $t$ (e.g., a well-known mineral spectrum). MF is a linear detector given by the formula: 
\begin{equation}
MF(x,t; \mu, \Sigma) = \frac{(t - \mu)^T \Sigma ^{-1} (x - \mu)}{(t - \mu)^T \Sigma ^{-1} (t - \mu)}.
\end{equation}

RX also uses scene statistics $\mu$ and $\Sigma$. However, it does not measure similarity between two spectra, but rather how anomalous a spectrum $x$ is with respect to the scene. Large values indicate outliers. Its computation is similar to MF:
\begin{equation}
RX(x; \mu, \Sigma) = (x - \mu)^T \Sigma ^{-1} (x - \mu).
\end{equation}

Finally, ongoing and future work consists of spectral unmixing using deep learning \cite{Candela2021}. 

\begin{figure*}[t!]
\centering
\includegraphics[width=\linewidth]{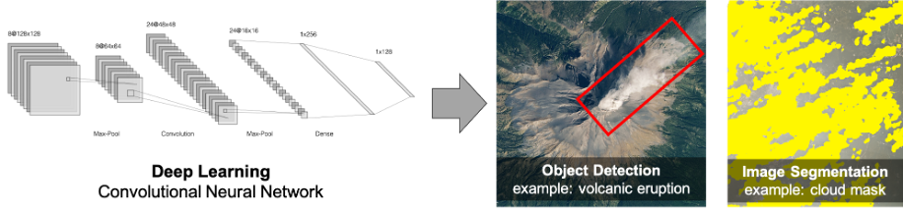}
\caption{CNNs will be used for the onboard analysis of many science events.}
\label{fig:ml}
\end{figure*}

\subsection{Convolutional Neural Networks}

Image analysis consists of semantic segmentation using the U-Net \cite{Ronneberger2015} deep CNN architecture tailored for deployment on flight hardware. We require models that provide high quality, quick classification using minimal computing resources. To minimize CPU computation, we embed preprocessing operations such as normalization as layers in the CNNs.

The trained models identify clouds, surface water extent (flooding), thermal events (e.g., volcanoes, wildfires), land surface type (e.g., city, forest, water, cropland,...), and harmful algal blooms. 

Models for segmentation applications, such as cloud screening and surface water extent, are trained to optimize the sparse categorical cross entropy loss. For thermal detection, positive classification is much rarer, therefore we optimize a weighted version of the sparse categorical cross entropy loss. The weights are set based on the distribution of classes in the training set.

\begin{table}[b!]
    \centering
    \begin{tabular}{c|c|c|c}
        Application & Clouds & SWE & Thermal \\ \hline
        Accuracy & 0.9748 & 0.8730 & 0.9988 \\
        Positive IoU & 0.9063 & 0.7069 & 0.9715 \\
        Negative IoU & 0.9144 & 0.8101 & 0.9988 \\
    \end{tabular}
    \caption{Results of current, best performing U-Net models on test set data for three applications. Models in the table perform binary classification, therefore positive refers to the label 1 and negative to label 0.}
    \label{tab:my_label}
\end{table}

\begin{table}[t!]
    \centering
    \resizebox{\columnwidth}{!}{
    \begin{tabular}{c|c|c|c}
        Application & Model & Model Size & Execution Time (s) \\ \hline\hline
        Clouds & U-Net Xception & 4.5 MB & 0.4781\\
         & U-Net UAVSAR & 4.3 MB & 0.5293  \\ \hline 
        SWE & U-Net Xception & 4.5 MB & 0.4938  \\
         & U-Net UAVSAR & 4.3 MB & 0.5333 \\ \hline 
        Thermal & U-Net Xception & 4.5 MB & 0.4800 \\
         & U-Net UAVSAR & 4.3 MB & 0.5307 \\ \hline 
        Vegetation & SAM & 175 KB & 2.499 \\
         & MF & 177 KB & 4.3060 \\
         & RX & 3 KB & 3.119 \\\hline 
        Mineral & SAM & 4 KB & 10.0871  \\
         & MF & 6 KB & 17.9930 \\
         & RX & 3 KB & 13.5023 \\ 
    \end{tabular}
    }
    \caption{Model size and single input execution time when compiled and executed on a Myriad X VPU.}
    \label{tab:exec_times}
\end{table}

\subsection{Evaluation and Verification}

We evaluate the quality of the algorithms on ground hardware and verify their computation on flight hardware prior to flight. Table \ref{tab:my_label} shows the accuracy and intersection over union (IoU) of three binary image classifiers on the test data sets. Cloud and thermal detection both achieve over $97\%$ accuracy and higher negative (label 0) IoU compared to positive (label 1) IoU. Surface water extent achieves lower performance than the other classifiers, but still maintains $87\%$ accuracy. It is desirable for the models to over-classify rather than have false negatives in many applications.

\begin{figure}[t!]

\centering
\includegraphics[width=1\columnwidth]{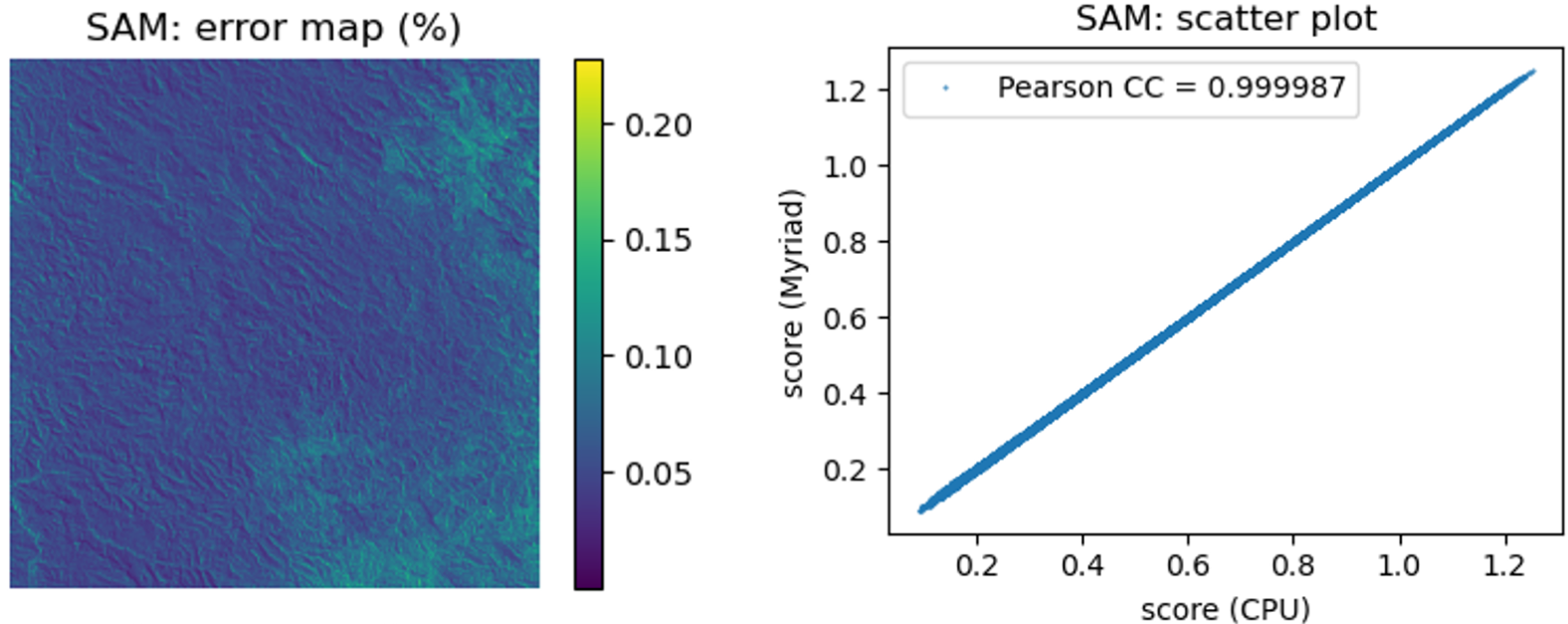}
\includegraphics[width=1\columnwidth]{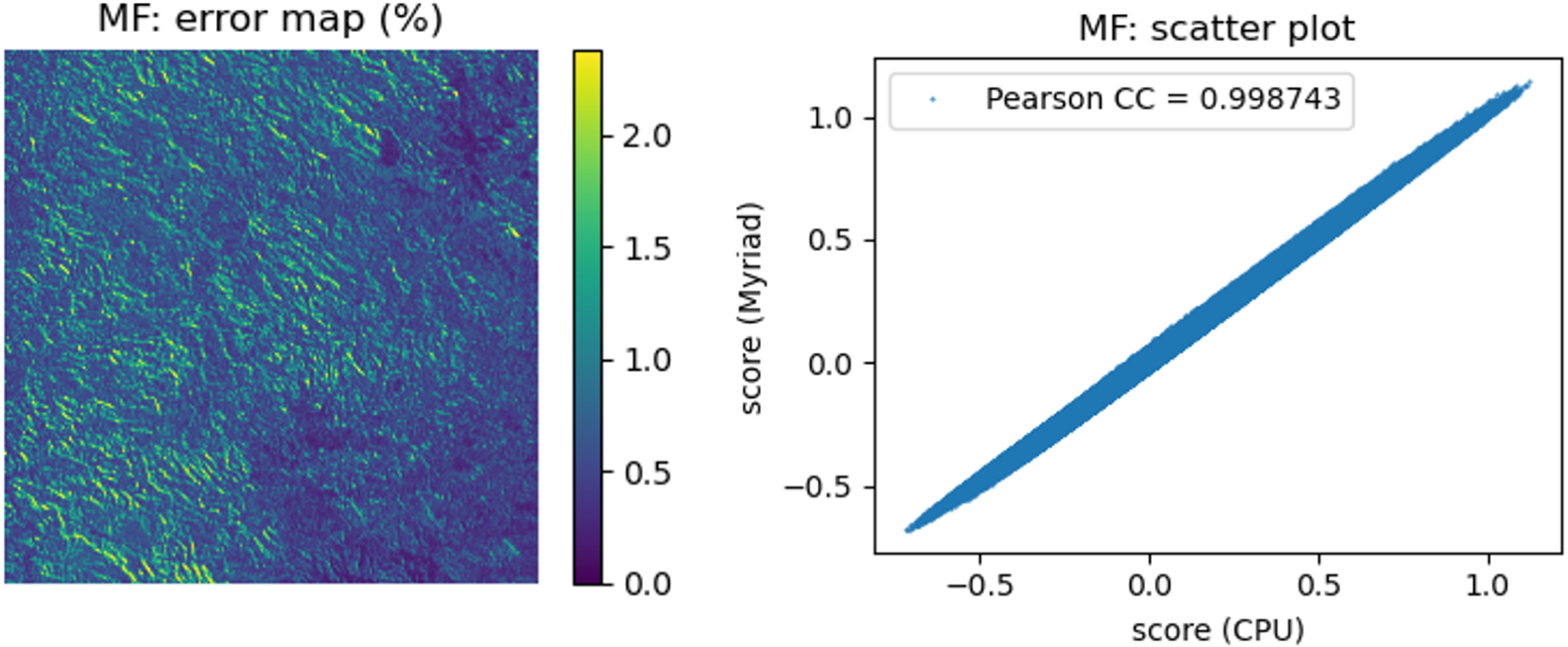}
\includegraphics[width=1\columnwidth]{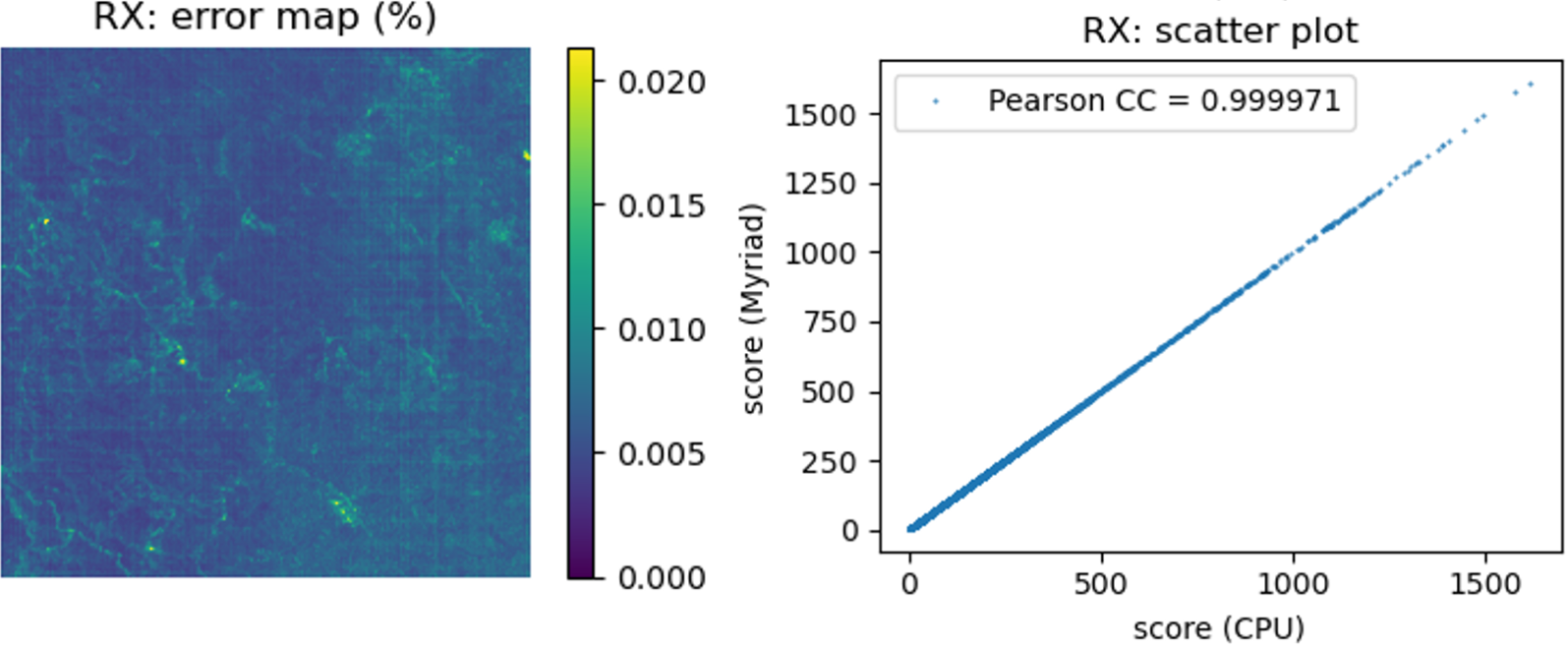}
\caption{Validation of spectral analysis algorithms computation on a CPU and Myriad X VPU.}
\label{fig:error}
\end{figure}

In addition to evaluating model quality, we verify model size, run-time and error on both CPU an Myriad X hardware. Table \ref{tab:exec_times} shows the model sizes and single input execution times when run on a Myriad X VPU. The inputs to the spectral algorithms are higher dimensional resulting in a longer runtime despite smaller model size compared to the U-Nets. Figure \ref{fig:error} illustrates the error of the spectral algorithms for vegetation detection when run on the Myriad X versus a traditional CPU. As confirmed by the analysis, we expect to see nearly identical outputs barring minor offsets due to floating point arithmetic.

\section{Current Status}

\begin{figure*}[t!]
\centering
   \includegraphics[width=\columnwidth]{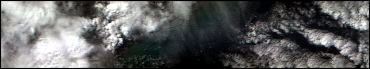}
    \includegraphics[width=\columnwidth]{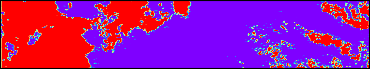}
   \includegraphics[width=\columnwidth]{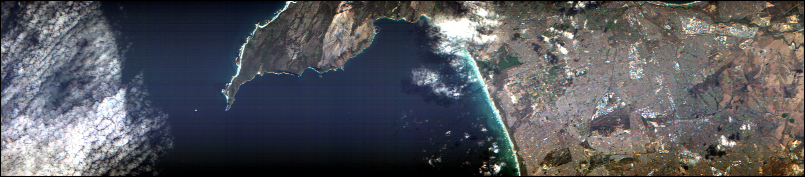}
   \includegraphics[width=\columnwidth]{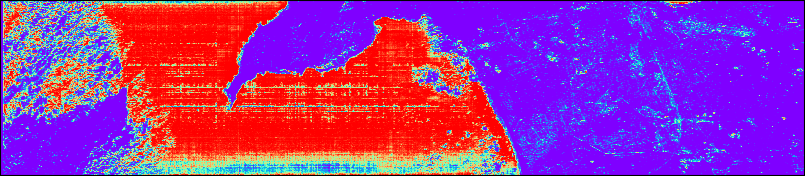}
\caption{Two scenes taken by CS-6 (left), and the respective inference that would be obtained onboard for cloud screening (top right) and surface water extent (bottom right) from deep learning models. \textit{Includes imagery from CogniSAT-6/HAMMER, 2024, Open Cosmos Limited. All rights reserved.}}
\label{fig:hyperscape-ml}
\end{figure*}

\begin{figure*}[t!]
\centering
\includegraphics[width=2\columnwidth]{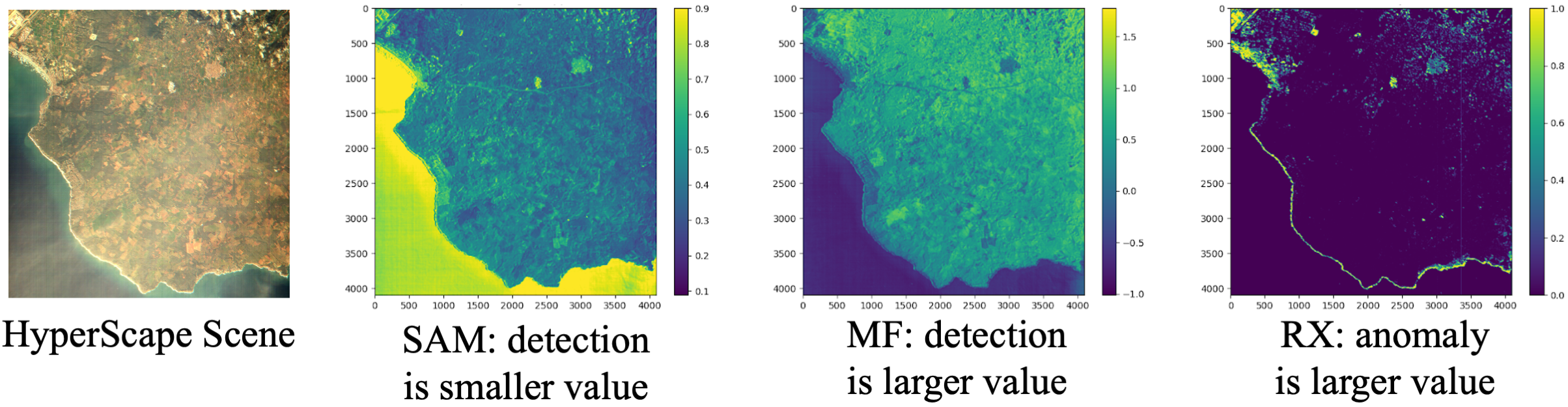}
\caption{The detection of vegetation for a CS-6 scene (left) using three spectral analysis algorithms. \textit{Includes imagery from CogniSAT-6/HAMMER, 2024, Open Cosmos Limited. All rights reserved}.}
\label{fig:spectral-hs}
\end{figure*}

Development begins with training of deep learning models and engineering of spectral algorithms. These models are then tested on Myriad X Neural Compute Stick(s). Finally, the models are executed on a flatsat testbed at Ubotica before upload and use onboard CS-6. 

We have developed dozens of models for the applications listed previously. An initial, small set of these models have been verified on the flatsat testbed and are pending flight on CS-6 in September of 2024. These models include CNNs for cloud screening, surface water extent, and thermal activity detection. Figure \ref{fig:hyperscape-ml} shows the classification of clouds and surface water extent for two scenes taken by CS-6 as would be computed onboard. The demonstration will include the collection of a scene, the onboard data inference of that collection, and the receipt of an ISL message containing a summary of the data. The complete data and segmentation will be received via a downlink after the experiment for further analysis.

\begin{figure}[!h]
\centering
\includegraphics[width=\columnwidth]{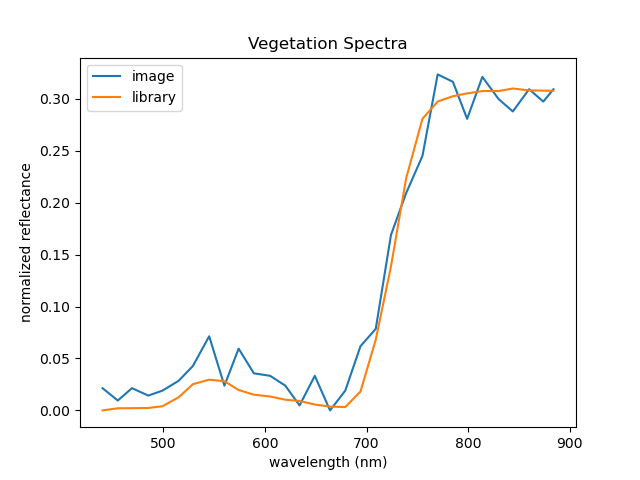}
\caption{Comparison of vegetation spectra: CS-6 imagery (blue) vs.\ one example from the USGS spectral library (orange).}
\label{fig:spectral-vegetation}
\end{figure}

The spectral analysis algorithms will be demonstrated onboard CS-6 after the CNNs in the fall of 2024. On ground hardware, the algorithms have been verified for mineral and vegetation detection applications with CS-6 imagery as shown in Figures \ref{fig:spectral-hs} and \ref{fig:spectral-vegetation}.

\section{Conclusion}
Leveraging edge computing for onboard data analysis is an exciting new capability of Earth-observing assets that opens the door for new Earth science. Spectral analysis can provide insights into high dimensional data. Image analysis can quickly detect features of interest in scenes. Engineering these processes to execute at the edge requires lightweight and efficient models that maintain high performance. We hope to advance the technology readiness level of this capability to enable deployment to future Earth-science missions.
In addition to demonstrations of more applications and models onboard CS-6, we have plans to deploy these models to more spacecraft. We also plan to integrate the onboard inference with other technologies such as \textit{dynamic targeting} and \textit{multi-asset federated scheduling} in future flight demonstrations \cite{nos-igarss,dt-isairas}. 

\section{Acknowledgments}

Portions of this work were performed by the Jet Propulsion Laboratory, California Institute of Technology, under a contract with the National Aeronautics and Space Administration (80NM0018D0004). This work was supported by the NASA Earth Science and Technology Office (ESTO). Government sponsorship acknowledged.

\printbibliography

\end{document}